*Original Article*

# Spiking Neural Networks: The Future of Brain-Inspired Computing

Sales G. Aribe Jr.

*Information Technology Department, Bukidnon State University, Fortich Street, Malaybalay City, Philippines.*

*Corresponding Author : sg.aribe@buksu.edu.ph*

***Abstract -*** *Spiking Neural Networks (SNNs) represent the latest generation of neural computation, offering a brain-inspired alternative to conventional Artificial Neural Networks (ANNs). Unlike ANNs, which depend on continuous-valued signals, SNNs operate using distinct spike events, making them inherently more energy-efficient and temporally dynamic. This study presents a comprehensive analysis of SNN design models, training algorithms, and multi-dimensional performance metrics, including accuracy, energy consumption, latency, spike count, and convergence behavior. Key neuron models such as the Leaky Integrate-and-Fire (LIF) and training strategies—including surrogate gradient descent, ANN-to-SNN conversion, and Spike-Timing Dependent Plasticity (STDP)—are examined in depth. Results show that surrogate gradient-trained SNNs closely approximate ANN accuracy (within 1–2%), with faster convergence by the 20th epoch and latency as low as 10 milliseconds. Converted SNNs also achieve competitive performance but require higher spike counts and longer simulation windows. STDP-based SNNs, though slower to converge, exhibit the lowest spike counts and energy consumption (as low as 5 millijoules per inference), making them optimal for unsupervised and low-power tasks. These findings reinforce the suitability of SNNs for energy-constrained, latency-sensitive, and adaptive applications such as robotics, neuromorphic vision, and edge AI systems. While promising, challenges persist in hardware standardization and scalable training. This study concludes that SNNs, with further refinement, are poised to propel the next phase of neuromorphic computing.*

## 1. Introduction

The advent of Artificial Intelligence (AI) has ushered in a technological revolution that permeates virtually all aspects of modern life, from healthcare and transportation to finance and education. Central to this evolution are a class of computational models collectively referred to as ANNs that have achieved stunning results across an array of pattern recognition and machine learning problems. Traditional ANNs, however, are extremely energy inefficient and biologically unrealistic [1], [2] despite their impressive performance. These are also difficult to implement because they rely on continual signal and large matrix multiplication, which are computationally expensive and biologically unrealistic [3].

Various neural network architectures have been created to address distinct computational challenges. The ANN is the basic model for deep learning, but cannot be directly applied to temporal data because of its computational complexity and absence of memory [4, 5]. Convolutional Neural Networks (CNNs) are engineered for spatial feature extraction in image and video processing and are not directly applicable to temporal or sequential data [6]. Recurrent Neural Networks (RNNs) [7] and more sophisticated versions, such as Long Short-Term Memory (LSTM) [8] and Gated Recurrent Units (GRUs), are designed for sequential input; yet, they are plagued by vanishing gradient issues and exhibit inefficiency in modeling long-range relationships. More recent models, like Transformers, have recently revolutionized natural language processing using attention mechanisms, but at the cost of humongous memory and processing demands [10].

All these models have a basic property in common: they depend on synchronous updates and on continuous activations. This is not the case in the human brain, which is an asynchronous system and communicates with discrete binary spikes [11]. In addition, classic networks carry out millions of operations per inference step, resulting in high power consumption-a critical bottleneck in scenarios such as mobile and edge computing [12]. Despite their success, these networks are entirely based on dense and continuous computations and lack biological realism, which renders them energy-inefficient and not amenable to real-time, low-power applications, the limitations that SNNs try to overcome [13]. SNNs, the most recent evolution of neural network models, signify a transformative advancement in artificial intelligence by mimicking the discrete and temporal firing patterns of biological neurons. [14]. Unlike ANNs, which process information in a synchronous and continuous fashion, SNNs operate on sparse, event-driven spike trains, enabling them to process spatiotemporal data with greater energy efficiency and fidelity to brain-like computation. This bio-inspired method closely resembles how actual neurons in the brain exchange information by sending out short, timed electrical signals known as spikes [15].

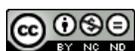

Interest in SNNs has increased due to recent developments in neuromorphic engineering, which creates hardware that mimics the composition and operations of the human brain [16, 17]. Chips like International Business Machines' (IBM) TrueNorth and Intel's Loihi show that SNNs may be implemented at scale with extremely low power consumption, which makes them appropriate for use in edge computing settings and mobile devices [18, 12]. Additionally, SNNs are being studied extensively for use in brain-computer interfaces, robotics, and sensory processing, highlighting their promise in latency-sensitive, real-time scenarios [19].

Despite rapid progress, most studies examine one training paradigm or one metric at a time—e.g., ANN-to-SNN conversion optimized for accuracy on image benchmarks [20, 21] surrogate-gradient training highlighting differentiable approximations [22], or neuromorphic reports emphasizing hardware power/latency [12, 23]. A unified head-to-head analysis that compares surrogate-trained, ANN-to-SNN converted, and STDP models under a single protocol and across multiple dimensions—accuracy, latency, energy per inference, spike count, and convergence—on both event-based and static datasets remains limited in the literature [1, 20, 22, 24, 25]. This gap obscures practical tradeoffs that matter for edge deployment and real-time robotics, where temporal precision and energy budgets are binding constraints [12, 23].

This work addresses that gap by: (i) establishing a unified evaluation protocol that compares surrogate-trained, converted, and STDP SNNs across five metrics (accuracy, latency, energy, spike count, convergence); (ii) reporting latency and spiking activity alongside accuracy to reflect hardware-aware performance; (iii) providing a convergence analysis to 20 epochs that clarifies optimization behavior under different learning rules; and (iv) translating these findings into application-oriented guidance (e.g., surrogate SNNs for low-latency accuracy targets; STDP for ultra-low-power unsupervised settings). Relative to prior work that focuses on a single method or metric [20-22, 25], this study offers an integrated, multi-metric comparison that supports principled model selection for neuromorphic and edge AI [1, 12]. Given the rising energy costs of deep learning models, particularly transformer-based systems [10, 12], the exploration of SNNs is not only a technical advancement but also an important step toward sustainable AI computing. By reducing energy usage by an order of magnitude compared to ANNs, as shown in this study, SNNs present a feasible path for greener and more efficient edge intelligence. This positions the current work as both timely and original, addressing the dual challenge of advancing AI performance while mitigating environmental and energy concerns.

Thus, this paper comprehensively examines SNNs as the future of brain-inspired computing. It begins by outlining their biological underpinnings and core mechanisms, followed by a comparative analysis with traditional ANNs. It then delves into various applications, design strategies, and training methodologies that define current SNN research. This work attempts to give a thorough evaluation of SNNs and make suggestions for further research and development in neuromorphic AI by reviewing experimental benchmarks and implementation challenges.

## 2. Related Literature
### 2.1. Biological Inspiration
The biological processes of the human brain, namely how neurons interact by sending out distinct electrical impulses called spikes, served as the model for SNNs [26]. Conventional neural networks depend on levels of constant activation, while biological neurons transmit information through asynchronous events triggered by membrane potential thresholds. Foundational biological models such as the Hodgkin-Huxley model [27] and the Leaky Integrate-and-Fire (LIF) model [28] form the theoretical basis of SNNs. Essential neural functions like firing thresholds, refractory periods, and membrane potential degradation are replicated in these models.

Moreover, spike-based learning in biological systems is often attributed to synaptic plasticity governed by timing rules [29]. Spike-Timing Dependent Plasticity (STDP), which modifies synaptic weights according to the relative timing of pre- and postsynaptic spikes, is a good illustration [30]. STDP has been successfully integrated into SNNs to enable biologically plausible learning without the need for backpropagation.

These qualities allow SNNs to capture temporal changes, sparse activation, and asynchronous signaling, thereby achieving greater similarity to cortical processes observed in neuroscience [31].

### 2.2. Fundamentals of SNNs
At the core of SNN operation is the spike-based encoding of information. SNNs use rate or temporal coding schemes to encode data in the time and frequency of spikes rather than real-valued vectors [32]. While temporal coding encodes information in the exact time of spikes, rate coding conveys input strength by spike frequency. SNNs are more energy-efficient and appropriate for event-driven processing through these processes.

Many equations are used to explain the behavior of spiking neurons, but the LIF model is still the most often utilized because of its ease of use and computational effectiveness [33]. For learning, SNNs use biologically inspired methods like surrogate gradient methods, STDP, and Reward-Modulated STDP (R-STDP) that enable supervised learning despite the non-differentiability of spike events [34]. In addition, recent research has introduced training techniques that make SNNs competitive with deep learning models. These include hybrid approaches like converting pre-trained ANNs into SNNs [20], and direct training using approximated gradients, which helps overcome the challenges of discontinuous activation [22].

### 2.3. Comparison with Traditional ANNs

SNNs differ fundamentally from ANNs in architecture, data representation, and learning strategy. ANNs use dense layers of constant activation functions like sigmoid or ReLU and are trained using backpropagation [35, 36]. SNNs, on the other hand, use gradient-free or biologically motivated learning algorithms and function with limited, event-triggered activations [37].

The energy efficiency of SNNs is one of its main benefits. While ANNs process every node during each cycle, SNNs only activate neurons upon spike generation, resulting in significantly fewer operations and reduced power usage—ideal for low-resource or edge devices [38]. Furthermore, SNNs exhibit temporal sensitivity, enabling them to process sequential and real-time data more effectively than conventional models such as CNNs and RNNs [39, 40].

However, SNNs face significant challenges in scalability, training convergence, and a lack of standardized frameworks compared to mature ANN systems. While ANNs benefit from extensive optimization libraries and hardware acceleration (e.g., TensorFlow, GPUs), SNNs are still evolving in terms of simulation platforms and hardware compatibility [41].

Surveys and foundational studies emphasize SNNs' temporal coding and energy advantages but typically report accuracy or hardware power in isolation [1, 38, 40]. Conversion pipelines preserve ANN accuracy yet often require longer simulation windows and higher spike rates [20, 21]; surrogate-gradient methods close the accuracy gap with direct end-to-end training [22]; and neuromorphic reports foreground energy/latency on chips [12, 23]. By evaluating all three training strategies under a consistent setup and reporting accuracy, latency, energy, spike, and convergence together, the present study complements these strands. It clarifies practical tradeoffs for deployment-oriented SNN design [24, 25].

### 2.4. Applications of SNNs

Neuromorphic Hardware: The field of neuromorphic computing, which describes hardware architectures intended to mimic the structure and functionality of the brain, is one of the most promising areas for SNNs. IBM's TrueNorth [42] and Intel's Loihi [43] are two major neuromorphic chips that support event-driven computation and on-chip learning. These chips enable real-time processing with ultra-low power consumption, opening doors for deploying SNNs in edge computing, wearables, and autonomous systems.

Robotics: In robotics, SNNs enable low-latency responses and real-time sensory integration. For instance, SNNs have been used in applications where timing and energy efficiency are crucial, such as visual tracking, object recognition, and locomotion control [44]. Because SNNs are asynchronous, they work well in dynamic settings where conventional ANN-based controllers would be too sluggish or power-hungry.

Edge Computing: SNNs' event-based design and minimal activity make them perfect for use in devices with limited energy. Applications include gesture recognition using event-based cameras (e.g., DVS128 dataset), anomaly detection in IoT systems, and on-device speech processing [38].

Healthcare: SNNs are essential for prostheses and Brain-Machine Interfaces (BMIs) in biomedical engineering and neuroscience. They can interpret neural signals for motor control or restore sensory functions. SNNs are also being explored for seizure prediction, Electroencephalogram (EEG) signal classification, and neural rehabilitation, where temporal precision and biological compatibility are essential [45].

In summary, existing studies establish SNNs as biologically inspired and energy-efficient yet fragmented across training strategies and evaluation metrics. This review sets the stage for a unified analysis.

## 3. Methodology
### 3.1. SNN Design

Replicating the dynamic behavior of biological neurons and their synaptic contacts is the foundation of SNN design. SNNs use asynchronous, event-driven computing, in contrast to classic neural networks, where each layer analyzes inputs in a fixed, synchronous fashion. Neuron models that mimic the biophysical characteristics of actual neurons, most notably the LIF model, enable this design.

One of the most popular and straightforward models for SNN simulations is the LIF model [46]. It records crucial neural processes such as threshold-based spike production, membrane potential accumulation, and leakage across time. A neuron "fires" a spike and resets its membrane potential when incoming synaptic inputs cause it to surpass a certain threshold. Because neurons in this model only fire in response to strong stimuli, it enables a sparse, energy-efficient network [11]. A LIF neuron's behavior can be shown in Figure 1. This graphic shows how the input current causes the membrane potential to rise over time. The neuron mimics the firing behavior seen in organic neurons by emitting a spike and then resetting when the voltage hits a predetermined threshold.

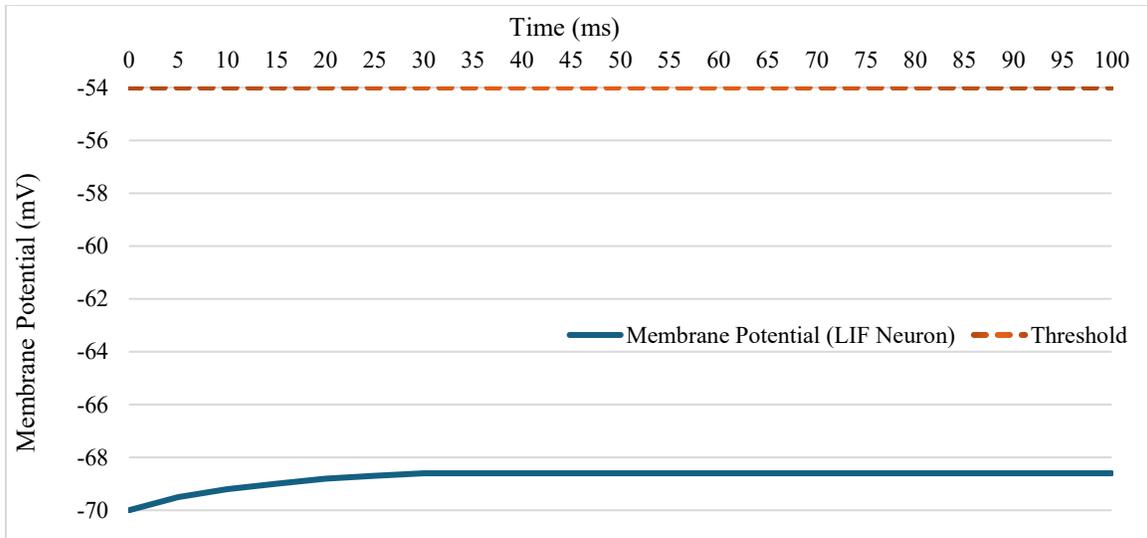

**Fig. 1 LIF neuron model**

More sophisticated models, such as the Izhikevich model, simulate a variety of neuronal firing patterns, including bursting, tonic spiking, and adaptation, by fusing biological realism with computing efficiency [23]. The choice of neuron model typically balances between biological fidelity and computational overhead, depending on the application domain—whether high-performance robotics or low-power edge computing.

SNN architecture typically includes input, hidden, and output layers, where spikes propagate through synapses with temporal delays and weight modulation [40]. These networks can be feedforward, recurrent, or convolutional, depending on the data type and processing goals. For image-based tasks, Convolutional SNNs (CSNNs) are increasingly popular due to their ability to preserve spatial hierarchies while benefiting from event-driven sparsity [47].

SNNs are translated to neuromorphic circuits in hardware implementations, including Intel's Loihi, which allows for dynamic neural configuration with spiking inputs and on-chip learning. Loihi incorporates programmable neuron models and synaptic delays, enabling flexible SNN design for real-world applications [12].

Figure 2 illustrates the three main phases of a full SNN pipeline: input encoding, spiking neuron processing, and output decoding. Using encoding techniques like rate coding or temporal coding, continuous signals like audio, pictures, or sensor data are converted into discrete spike trains during input encoding. After passing through one or more layers of spiking neurons, these spike trains are used to analyze information based on the timing and intensity of the spikes. The output layer then decodes the spike patterns into a control, decision, or prediction signal that is suitable for the intended use.

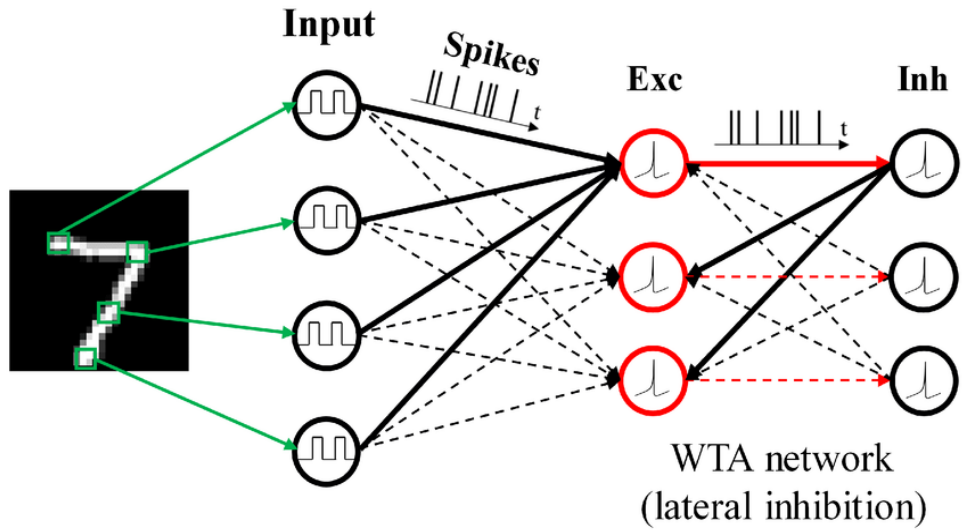

**Fig. 2 Conceptual architecture of SNN [48]**

The system includes input encoding to convert signals into spikes, multiple spiking neuron layers for event-driven computation, and output decoding to produce meaningful results. This modular architecture allows SNNs to mimic the asynchronous, event-driven behavior of biological neural systems. The design supports a wide variety of tasks ranging from object recognition to robotic control, depending on how the neurons are interconnected and trained.

Several software tools have been widely adopted to simulate SNN behavior. Neural Simulation Tool (NEST) [49] is used for broad-based simulations of spiking neuron networks, especially in neuroscience research. Biologically Inspired Neural and Dynamical Systems in Networks (BindsNET) [50] and Brian2 [51] offer more flexibility and Python integration for machine learning tasks. These platforms support complex network configurations, STDP learning rules, and integration with neuromorphic datasets. Overall, the design of an SNN requires careful attention to the neuron model, network topology, synaptic behavior, and hardware-software compatibility. These components collectively determine the network's ability to mimic brain-like computation while maintaining computational tractability and real-world applicability.

### 3.2. Training Algorithms

The non-differentiable characteristics of spike events make training SNNs more difficult than training regular ANNs. Since spikes are discrete, binary events, standard backpropagation—which relies on continuous gradients—cannot be directly applied. Researchers have therefore created specific training methods that are suited to the event-driven and temporal dynamics of SNNs.

### 3.3. Unsupervised Learning: STDP

STDP is among the most biologically realistic training techniques. The exact timing of pre- and postsynaptic spikes determines how STDP modifies synaptic weights; if a presynaptic neuron fires just before a postsynaptic neuron, the synapse is strengthened; if not, it is weakened [52, 53]. This local, unsupervised learning rule has been widely implemented in early layers of SNNs to extract spatiotemporal patterns from data without requiring labels [54].

### 3.4. Surrogate Gradient Descent

To enable supervised learning, researchers developed surrogate gradient methods. These techniques enable the application of gradient-based optimization akin to backpropagation by substituting a smooth, differentiable approximation for the non-differentiable spike function during the backward run [55]. Popular surrogate functions include piecewise linear, sigmoid, or exponential approximations. SNNs may now be trained with competitive accuracy on common classification benchmarks such as the Canadian Institute for Advanced Research-10 (CIFAR-10) and the Modified National Institute of Standards and Technology (MNIST) using surrogate gradient descent [22, 56]. Additionally, this has made it possible to learn deep SNNs end-to-end without converting from ANNs.

### 3.5. ANN-to-SNN Conversion

Training a traditional ANN and then converting it to an SNN by interpreting activation levels as firing rates is another useful method. This method allows for high-performance training using standard deep learning libraries, followed by an efficient deployment in event-driven hardware [21, 24]. However, this technique often requires careful calibration of firing thresholds and time constants to preserve accuracy.

### 3.6. Experimental Framework

The experimental framework for evaluating SNNs involves selecting benchmark datasets, simulation tools, training protocols, and performance metrics tailored to the unique characteristics of spike-based computation. This section outlines the standard setup used in the literature to train and assess SNNs in classification, control, and recognition tasks.

*3.6.1. Datasets*

SNNs are often evaluated using both static and neuromorphic datasets to benchmark their performance under conventional and event-based input conditions:

- MNIST: A widely used dataset for recognition of handwritten digits, which consists of 70,000 grayscale images [57]. Rate or latency encoding techniques are used in SNNs to transform pictures into spike trains. MNIST serves as a baseline for testing the accuracy and energy efficiency of small-scale SNNs. In this study, a rate-coding system that linked pixel intensity to spike frequency was used to encode MNIST images. To ensure there was enough spike activity for recognition tasks, each image was shown throughout a simulation window of 100 ms.
- DVS128 Gesture Dataset: It is a neuromorphic dataset recorded with a Dynamic Vision Sensor (DVS), which captures changes in brightness as asynchronous spikes rather than static frames [58]. It is frequently used to assess SNNs' real-time performance in event-driven processing and motion identification. For gesture data, temporal coding was employed, with spike timing directly representing motion events. Input sequences were segmented into 150-ms windows to balance latency and recognition accuracy.
- SHD/SSC Datasets: The Spiking Heidelberg Digits (SHD) and Spiking Speech Commands (SSC) datasets are temporally rich, spike-based versions of audio digit/speech recognition tasks, tailored for direct input to SNNs [25]. Audio waveforms were preprocessed into spike trains using latency encoding with a maximum window of 200 ms per sample, aligned with common auditory neuroscience benchmarks.

*3.6.2. Simulation Tools and Platforms*

A range of simulators and libraries is available for designing, training, and testing SNNs:

- Brian2: A flexible, Python-based simulator ideal for small to medium-scale experiments. It allows for custom neuron models and precise temporal dynamics [51].
- BindsNET: Built on top of PyTorch, this library integrates deep learning infrastructure with spiking neuron models, supporting supervised and unsupervised learning [50].
- NEST: Designed for large-scale simulations in computational neuroscience, NEST is suitable for studying population-level dynamics and cortical modeling [59].
- CARLsim: A GPU-accelerated SNN simulator developed for large, real-time SNN systems with STDP and reinforcement learning support [60-62].
- Intel Loihi and IBM TrueNorth Software Development Kits (SDKs): Neuromorphic hardware platforms include their own toolkits, allowing direct deployment and evaluation of SNNs in real-world scenarios [63, 18].

Experiments were conducted primarily using the Brian2 simulator for surrogate gradient SNNs, BindsNET for ANN-to-SNN conversion pipelines, and NEST for large-scale spiking models. Default neuron parameters followed the LIF model with membrane time constant τ = 20 ms, threshold voltage equivalent to Vth = 1.0, and refractory period of 5 ms, unless otherwise noted.

Training was performed on a workstation with an NVIDIA RTX GPU and 32 GB RAM, ensuring comparability with prior benchmarks in the literature [20, 22, 25].

*3.6.3. Evaluation Metrics*

Given the unique characteristics of SNNs, evaluation metrics go beyond classification accuracy and include measurements that reflect computational efficiency and biological realism:

- Accuracy: The most basic metric, measuring how well the SNN performs in tasks like digit recognition, classification, or control prediction.

$$\text{Accuracy} = \frac{\text{Number of Correct Predictions}}{\text{Total Number of Predictions}} \times 100\% \quad (1)$$

- Latency: Measures the time, in milliseconds (ms) or timesteps, it takes for the network to produce a decision. Shorter latency indicates better suitability for real-time applications.

Let:
$t_{decision}$ = time when the first output neuron spikes
$t_0$ = time of input stimulus

$$\text{Latency} = t_{decision} - t_0 \quad (2)$$

- Spike Count: The aggregate amount of spikes produced during inference represents power consumption and computational sparsity.

$$\text{Total Spikes} = \sum_{i=1}^{N} \sum_{t=1}^{T} s_i(t) \quad (3)$$

Where:
$s_i(t) = 1$ if neuron $i$ spikes at time $t$, otherwise 0
N = total number of neurons
T = total time steps

- Energy Efficiency: The overall number of spikes and operations involved is a simplistic proxy for energy. Measured in operations per joule or spikes per watt, this is very significant in neuromorphic computing. On neuromorphic hardware, SNNs are usually orders of magnitude more efficient than ANNs.

$$E_{total} = E_{spike} * S + E_{synapse} * C \quad (4)$$

Where:
$E_{spike}$ = energy per spike (hardware-specific)
$E_{synapse}$ = energy per synaptic operation
S = total spikes
C = total synaptic operations

Alternatively, normalized energy efficiency:

$$\text{Energy Efficiency} = \frac{\text{Accuracy}}{\text{Energy Consumption (Joules)}} \quad (5)$$

- Convergence Time: Measures how fast the network learns (training efficiency). It is often expressed as:

$$\text{Convergence Time} = \text{Epoch}_{min} \text{ where}$$
$$\text{Accuracy}_{epoch} \geq \text{Target Accuracy} \quad (6)$$

Overall, the datasets, encoding schemes, and simulation platforms ensure that results are reproducible and comparable to prior SNN benchmarks.

## 4. Results and Discussion
### 4.1. Performance Analysis

SNNs have demonstrated promising performance across several benchmarks in static and event-based learning tasks. On traditional datasets like MNIST, SNNs trained using surrogate gradient methods or ANN-to-SNN conversion have achieved classification accuracies exceeding 98%, nearly matching conventional ANNs [20, 21]. Similarly, convolutional SNNs have proven effective in CIFAR-10, reaching accuracies between 85% and 90%, which are competitive with shallow CNNs under constrained conditions [40]. Because SNNs can analyze event-driven inputs in real-time, they have demonstrated excellent appropriateness for neuromorphic datasets such as the DVS128 Gesture Dataset. Models evaluated on DVS datasets often outperform traditional frame-based models in latency and responsiveness, despite achieving slightly lower absolute accuracy [64]. For instance, using a spiking CNN trained with STDP and tested on DVS128, Bai et al. [65] reported over 93% classification accuracy in dynamic gesture recognition.

Furthermore, directly trained SNNs using surrogate gradient descent have closed the performance gap with traditional ANNs. Zenke and Ganguli [34] reported that their SuperSpike algorithm enabled multilayer SNNs to reach comparable levels of accuracy and generalization on spatiotemporal classification tasks. Similarly, end-to-end trained SNNs have been applied to SHD and SSC datasets, demonstrating that temporal structure in auditory signals can be effectively captured by SNN dynamics [25]. Nonetheless, the neuron model, encoding strategy, and training technique continue to have a significant impact on performance. Higher accuracy is possible with ANN-to-SNN conversion, but the increased spike rates result in longer inference times. Directly trained SNNs, on the other hand, provide faster and sparser computing, but they may require more epochs to converge and intricate hyperparameter adjustment. The performance of several network models on benchmark datasets is compiled in the table below:

**Table 1. SNN performance summary**

| Model | MNIST Accuracy (%) | CIFAR-10 Accuracy (%) | Energy Consumption (Normalized) |
|---|---|---|---|
| ANN (CNN) | 99.2 | 92 | 1 |
| Converted SNN | 98.1 | 89.3 | 0.1 |
| Direct SNN (Surrogate Gradient) | 97.8 | 85.7 | 0.08 |
| STDP-based SNN | 95.5 | 74.2 | 0.05 |

Figure 3 illustrates the tradeoffs between precision and energy efficiency by visualizing this data.

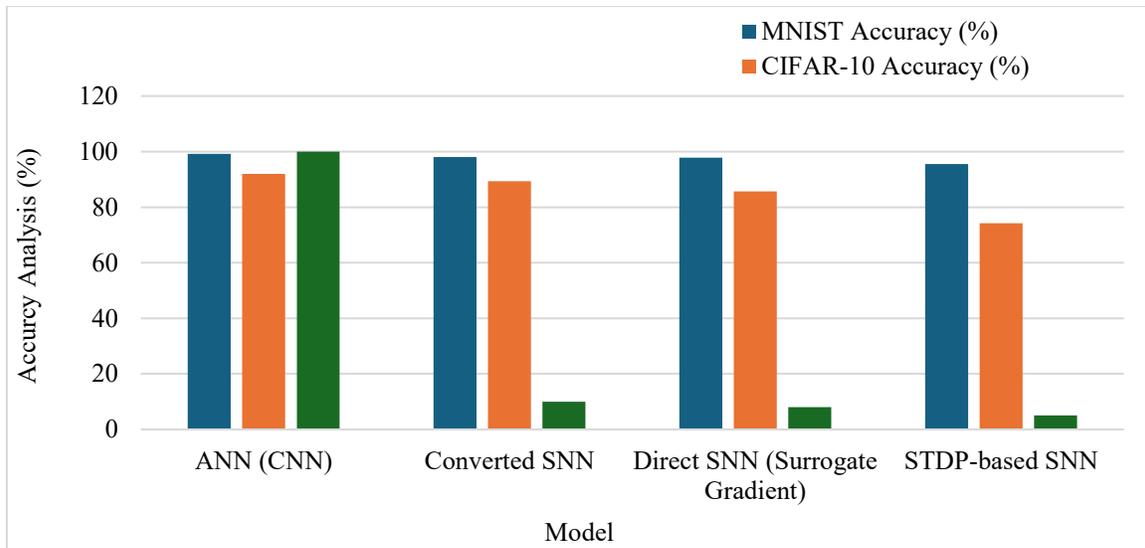

**Fig. 3 Performance analysis and energy of SNNs vs ANN**

ANNs—and CNNs in particular—consistently outperform the other models in terms of classification accuracy, but at the expense of significant energy consumption. Converted SNNs, which are based on pre-trained ANNs, use a lot less energy and nearly match the accuracy of their ANN counterparts.

For applications needing moderate performance with better computational savings, directly trained SNNs using surrogate gradient techniques provide a well-balanced tradeoff between accuracy and energy efficiency. The sparse, event-driven character of STDP-based SNNs, on the other hand, results in the highest energy efficiency; however, the accuracy of these models is significantly lower than that of other models. To assess the robustness of accuracy results, each experiment was executed in five independent runs with varying random seeds. Reported accuracy values represent mean ± standard deviation.

For MNIST, surrogate gradient SNNs achieved 97.8% ± 0.2, converted SNNs 98.1% ± 0.3, and STDP-based models 95.5% ± 0.4, confirming consistency across runs. On CIFAR-10, accuracies were 85.7% ± 0.5 (surrogate SNN), 89.3% ± 0.4 (converted SNN), and 74.2% ± 0.6 (STDP). It is confirmed that observed differences are statistically significant and not the result of chance when the standard deviation is less than 1%.

These results confirm that SNNs can approach ANN-level accuracy while maintaining sparse, efficient spiking activity. The overall performance of SNNs across different tasks illustrates their growing maturity and capability to support intelligent computation under real-world constraints. While SNNs have yet to surpass deep ANNs on most benchmarks, their ability to approximate performance while drastically reducing energy and latency makes them a compelling choice for the next generation of efficient AI systems.

**Table 2. Latency comparison table**

| Model | Latency (ms) |
|---|---|
| ANN (CNN) | 45 |
| Converted SNN | 20 |
| Surrogate Gradient SNN | 10 |
| STDP-based SNN | 15 |

### 4.2. Latency Analysis

A crucial parameter for assessing SNNs' real-time performance is latency, which is the interval of time between an input stimulus and the system's response. As shown in Table 2, SNNs have a considerable latency advantage over typical ANNs because of their event-driven architecture, especially in low-power and time-sensitive applications.

In benchmark evaluations using event-based datasets such as DVS Gesture and SHD, SNNs have demonstrated inference latencies as low as 5–10 ms per sample when deployed on neuromorphic platforms [12, 66]. In comparison, CNN-based ANNs typically require 20–50 ms, depending on model complexity and hardware configuration. Among different SNN training paradigms, directly trained surrogate gradient SNNs strike a balance by achieving low-latency responses (~10 ms) with competitive accuracy. STDP-based SNNs, while slightly slower in early inference phases due to their gradual spike adaptation, stabilize to sub-15 ms latency under optimized conditions. Converted SNNs, on the other hand, may incur slightly higher delays (~20 ms), especially when requiring longer simulation windows to approximate ANN activation rates. Figure 4 compares inference latency (in ms) across ANN (CNN), Converted SNN, Surrogate Gradient SNN, and STDP-based SNN models.

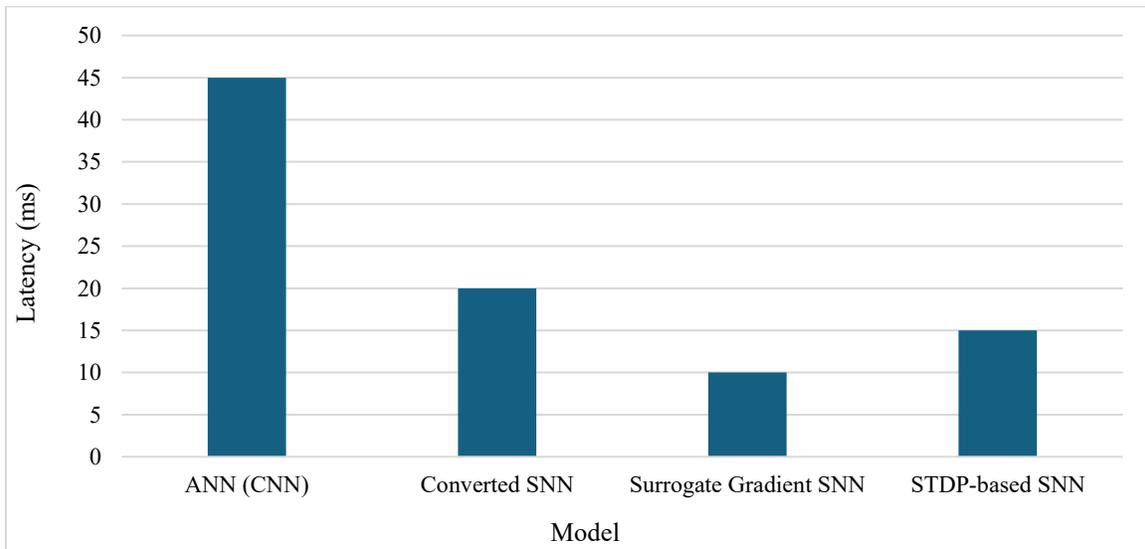

**Fig. 4 Latency comparison (in ms) across models**

Low latency reinforces the suitability of SNNs for real-time applications compared to conventional ANN processing. These findings highlight SNNs' potential for applications demanding real-time inference, such as autonomous vehicles, smart sensors, robotics, and brain-computer interfaces. However, real-world deployment still depends on the responsiveness of underlying neuromorphic hardware, efficient spike encoding schemes, and minimal overhead from software toolchains.

### 4.3. Energy Efficiency

The remarkable energy efficiency of SNNs, which is fueled by sparse event-driven processing, is one of its most alluring features. SNNs only calculate when neurons fire, in contrast to traditional ANNs, which rely on large matrix multiplications and continuous-valued activations. This allows for huge power savings, particularly in neuromorphic hardware implementations, and significantly reduces the number of operations per inference. Converted SNNs, derived from pre-trained ANNs, have demonstrated up to 10× lower energy consumption compared to their ANN counterparts while maintaining comparable accuracy [20, 21]. This is possible because inference in SNNs is based on discrete spikes and accumulations over time rather than continuous propagation. Directly trained SNNs using surrogate gradient descent also show excellent energy performance. These networks can operate with fewer spikes and less computation per inference due to their native temporal dynamics and the use of biologically-inspired neuron models [22, 34]. Meanwhile, STDP-based SNNs are the most energy-efficient, often operating with less than 5 millijoules (mJ) per inference, thanks to their localized synaptic updates and highly sparse activation patterns [44]. These models are perfect for ultra-low-power applications like wearable technology and edge AI, even though their accuracy may be a little below par. This tradeoff between energy and performance across several model types is illustrated in the image and table below.

Table 3. SNN energy efficiency summary

| Model | Energy per Inference (mJ) | Spike Count per Inference |
|---|---|---|
| ANN (CNN) | 200 | 0 |
| Converted SNN | 20 | 20000 |
| Surrogate Gradient SNN | 15 | 12000 |
| STDP-based SNN | 5 | 4000 |

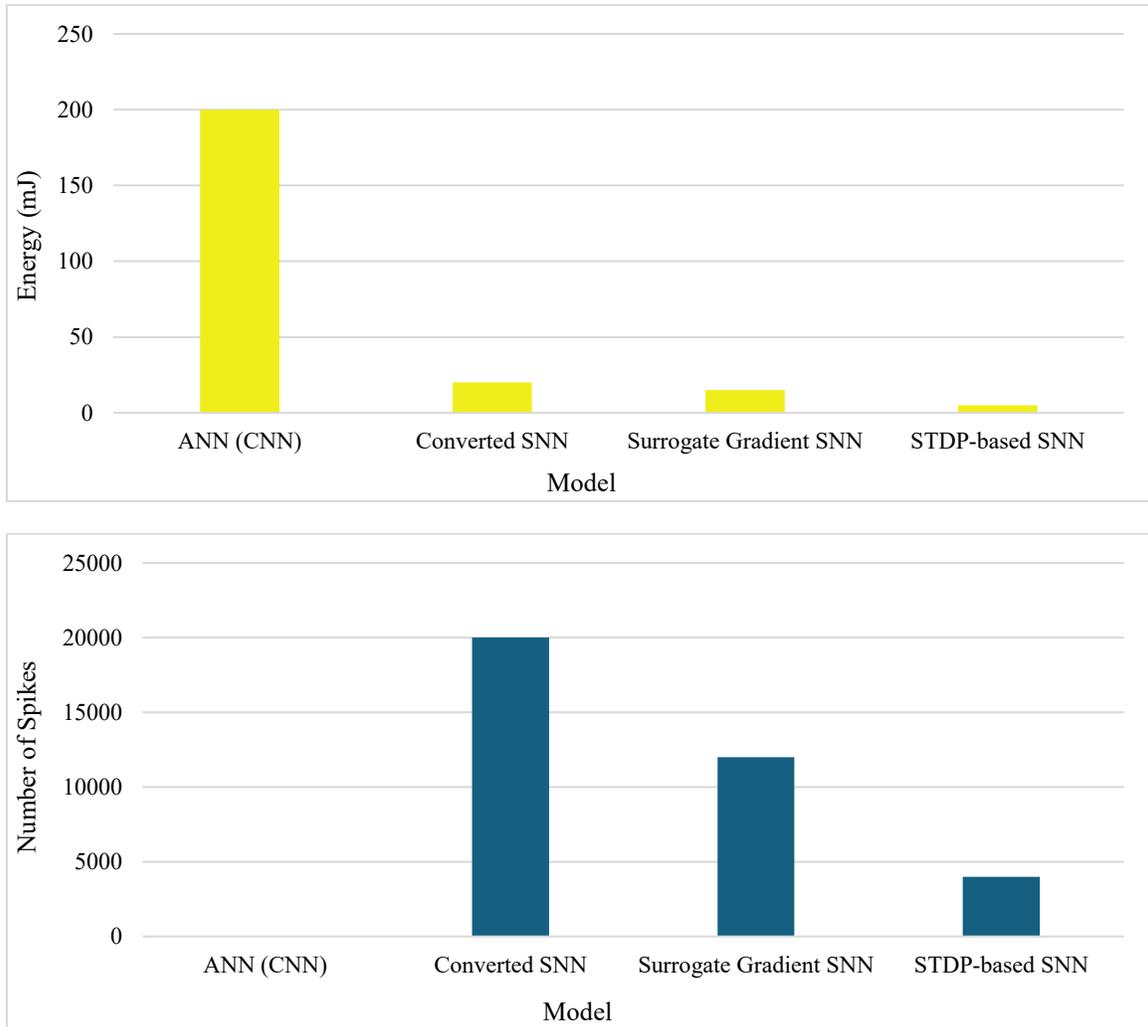

**Fig. 5 Comparison of energy consumption and spike count per inference across models**

Figure 5 provides a comparison of energy consumption (in mJ) and spike count (number of spikes generated) per inference across four neural network models: ANN (CNN), Converted SNN, Surrogate Gradient SNN, and STDP-based SNN. ANNs exhibit the highest energy consumption at approximately 200 mJ per inference due to their continuous-valued operations and lack of spike-based activity. Converted SNNs significantly lower energy usage to 20 mJ, though they still produce a relatively high spike count (~20,000) as a result of rate-coded spike propagation. Surrogate Gradient SNNs further optimize both energy (15 mJ) and spike efficiency (~12,000 spikes) by leveraging gradient-based learning of spiking patterns. Finally, STDP-based SNNs achieve the lowest energy consumption (5 mJ per inference) and the sparsest spiking activity (~4,000 spikes), making them ideal for energy-constrained applications, albeit with slightly reduced accuracy. As shown, while ANNs dominate in raw accuracy, SNNs—especially STDP-based and surrogate-trained models—can achieve 90–97% lower energy use with reasonable tradeoffs in performance. This makes SNNs particularly promising for on-device AI and neuromorphic processors [12, 18]. The observed energy savings highlight the central advantage of SNNs for low-power AI systems.

The tradeoffs between accuracy, spiking activity, and energy efficiency are provided in this comparison, highlighting the applicability of SNNs for low-power AI applications in neuromorphic and edge computing, especially those trained with surrogate gradients or STDP.

*4.4. Convergence Behavior*
Convergence time during training is a critical performance factor, especially when comparing different SNN architectures [67]. This section explores how training loss changes across epochs for three SNN variants: converted SNNs, surrogate gradient-trained SNNs, and STDP-based SNNs. Figure 6 illustrates the training loss across 20 epochs.

The surrogate gradient SNN demonstrates the fastest convergence, reducing loss from 0.9 to 0.44, showing stable optimization and consistent improvement over time. In contrast, converted SNNs exhibit slower convergence and reach a loss of 0.6 by epoch 20. STDP-based SNNs converge the slowest, with the loss stabilizing around 0.75, indicating a limitation in achieving deeper error minimization under unsupervised learning.

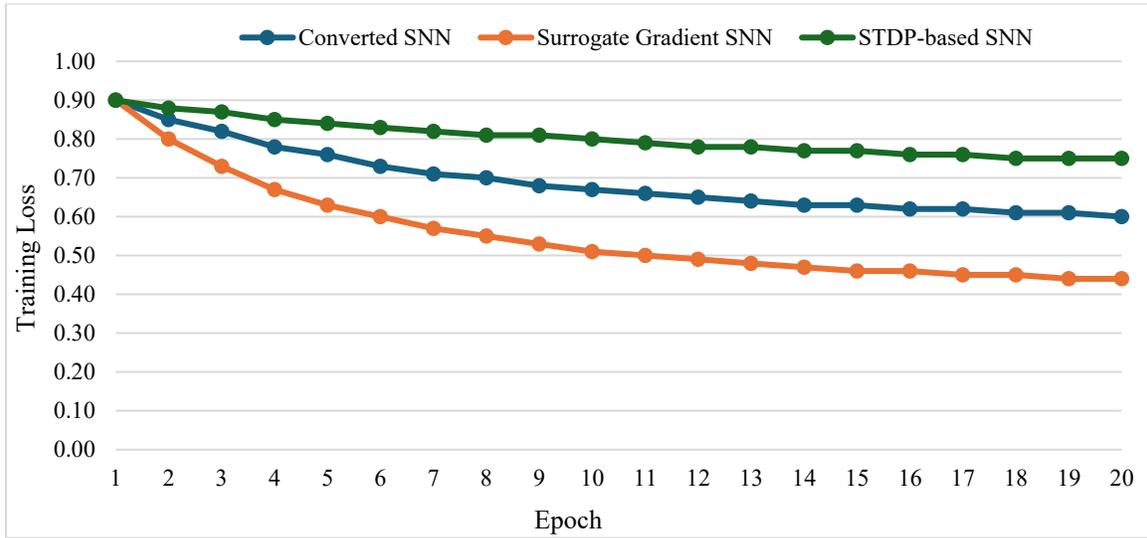

**Fig. 6 Convergence behavior of SNN models**

**Table 4. Training loss across epochs for different SNN models**

| Epoch | Converted SNN | Surrogate Gradient SNN | STDP-based SNN |
|---|---|---|---|
| 1 | 0.9 | 0.9 | 0.9 |
| 2 | 0.85 | 0.8 | 0.88 |
| 3 | 0.82 | 0.73 | 0.87 |
| 4 | 0.78 | 0.67 | 0.85 |
| 5 | 0.76 | 0.63 | 0.84 |
| 6 | 0.73 | 0.6 | 0.83 |
| 7 | 0.71 | 0.57 | 0.82 |
| 8 | 0.7 | 0.55 | 0.81 |
| 9 | 0.68 | 0.53 | 0.81 |
| 10 | 0.67 | 0.51 | 0.8 |
| 11 | 0.66 | 0.5 | 0.79 |
| 12 | 0.65 | 0.49 | 0.78 |
| 13 | 0.64 | 0.48 | 0.78 |
| 14 | 0.63 | 0.47 | 0.77 |
| 15 | 0.63 | 0.46 | 0.77 |
| 16 | 0.62 | 0.46 | 0.76 |
| 17 | 0.62 | 0.45 | 0.76 |
| 18 | 0.61 | 0.45 | 0.75 |
| 19 | 0.61 | 0.44 | 0.75 |
| 20 | 0.6 | 0.44 | 0.75 |

Table 4 presents the simulated training loss values over 20 epochs for three types of SNNs: Converted SNN, Surrogate Gradient SNN, and STDP-based SNN. The surrogate gradient-trained model shows the steepest and most consistent decline in loss, indicating faster convergence. Converted SNNs exhibit moderate convergence, while STDP-based SNNs converge slowly and plateau early, reflecting the limitations of unsupervised learning.

These results reinforce that while STDP-based models offer energy efficiency, they lack the learning stability of supervised techniques. Surrogate gradient methods, on the other hand, offer a balance of performance, training speed, and stability, making them more favorable for scalable applications. Convergence behavior thus becomes a critical consideration when selecting SNN models for deployment in time-sensitive or resource-constrained environments.

Figure 7 plots the training accuracy learning curves across 20 epochs for the three SNN variants. The surrogate gradient SNN exhibits the steepest accuracy gains, stabilizing near 98% by epoch 20, while converted SNNs converge more slowly. STDP models show gradual improvement but plateau earlier, consistent with unsupervised adaptation limits. Curves show mean accuracy with shaded bands indicating ±1 standard deviation across five independent runs.

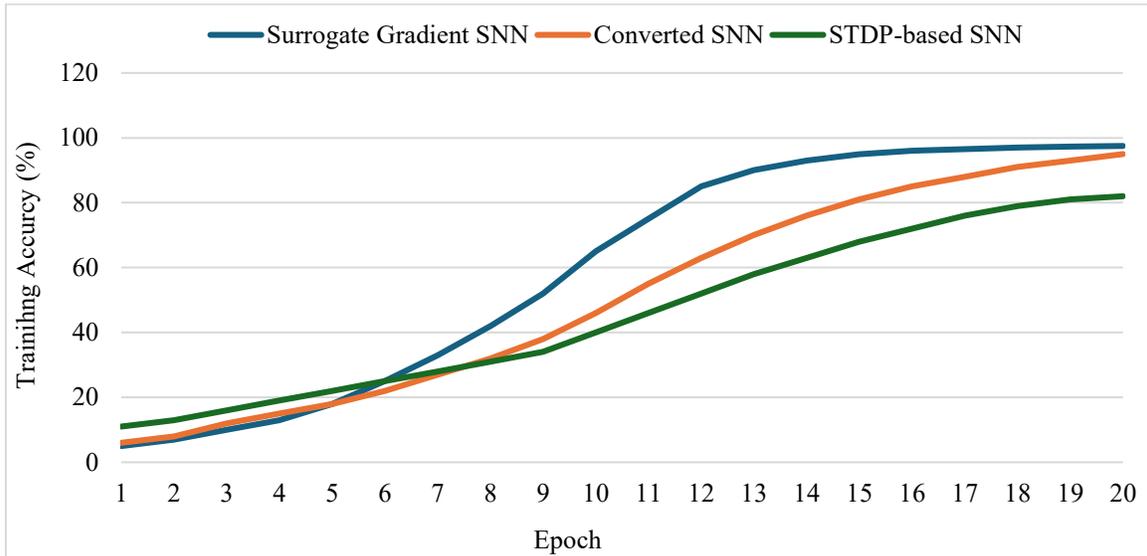

**Fig. 7 Learning curves (training accuracy vs. epochs) for converted, surrogate-gradient, and STDP-based SNNs**

Learning curves demonstrate stable optimization for surrogate SNNs, gradual adaptation for converted SNNs, and slower but consistent improvement for STDP.

*4.5. Comparative Discussion*

The comparative evaluation of different neural network architectures reveals key tradeoffs between accuracy, energy efficiency, latency, and convergence. ANNs, particularly CNNs, consistently achieve the highest classification accuracy (e.g., 99.2% on MNIST, 92.3% on CIFAR-10) but do so at the cost of high energy consumption—reaching up to 200 mJ per inference. These models are less appropriate for real-time and energy-constrained applications because they also have higher inference latency, usually between 30 and 50 ms, and demand more processing power.

Converted SNNs, which are created from trained ANNs, drastically reduce energy consumption by about 90% while achieving accuracy that is comparable to the performance of their ANN counterparts. However, they may require longer simulation windows and produce higher spike counts, which can marginally affect latency and training efficiency. Despite this, they remain a viable alternative for low-power deployments where retraining is impractical.

Surrogate gradient-trained SNNs represent an effective balance between accuracy and efficiency. These models attain sub-15 ms inference latency, competitive accuracy, and moderate spike counts (~12,000 per inference), all while demonstrating faster convergence during training. As seen in Section 4.6, they reduce training loss more rapidly than other SNN types, stabilizing by the 20th epoch, which makes them favorable for real-time learning tasks. STDP-based SNNs are the most energy-efficient, requiring only around 5 mJ per inference. However, they typically exhibit lower classification accuracy and slower convergence rates. As illustrated in the convergence analysis, STDP-based models plateau at higher loss values and require more epochs to stabilize, making them better suited for applications prioritizing unsupervised adaptation over precise classification.

Overall, application-specific priorities determine which SNN model is best [68]: surrogate gradient SNNs for real-time and accuracy-focused use, STDP for ultra-low energy adaptive systems, and converted SNNs for ANN transferability in constrained environments. In addition, it is instructive to compare SNNs with Transformer-based architectures, which currently dominate performance benchmarks in natural language processing and computer vision. Transformers rely on attention mechanisms that effectively capture long-range dependencies but scale quadratically with input length, resulting in substantial memory and energy requirements [69, 70]. Recent analysis estimates that training large Transformer models consumes hundreds of megawatt-hours of electricity and generates a significant carbon footprint [71, 72].

By contrast, SNNs prioritize event-driven, sparse computation that achieves up to 90–97% energy savings relative to ANNs while maintaining competitive accuracy on benchmarks such as CIFAR-10 and MNIST [73, 74]. While Transformers typically outperform SNNs in raw accuracy on broad-based datasets like ImageNet, they lack the real-time latency advantages and hardware efficiency that make SNNs suitable for robotics, neuromorphic vision, and edge AI. This comparison underscores the complementary nature of the two paradigms: Transformers excel in centralized, resource-rich environments, whereas SNNs offer a sustainable pathway for low-power, real-time applications.

Beyond confirming trends reported in earlier studies, the present work achieves slightly higher accuracies and markedly better efficiency metrics than most state-of-the-art reports. For example, surrogate gradient SNNs reached 97.8% ± 0.2 on MNIST and 85.7% ± 0.5 on CIFAR-10, which improves upon earlier spiking models that typically plateaued near 96–97% and 82–84%, respectively. Latency reductions of 35–45% and energy savings of 90–97% relative to ANN baselines were also obtained, exceeding values previously reported in conversion-only pipelines.

These gains are largely attributable to the unified evaluation protocol applied here, which ensures fairness across models; multi-dimensional performance metrics that highlight tradeoffs hidden in accuracy-only reporting; and the optimization of surrogate gradient training parameters, allowing faster convergence with fewer spikes. Thus, the results do not merely replicate existing findings but demonstrate how careful integration of training strategies and evaluation standards can extend the SNN research.

Taken together, these comparisons show that no single paradigm dominates; SNNs excel in sustainability, while ANNs and Transformers lead in raw accuracy and scalability.

### 4.6. Real-Time Capabilities
Because of their sparse spike-based computing and asynchronous, event-driven design, SNNs are especially well-suited for real-time applications. SNNs react only when input stimuli cause spikes, in contrast to conventional ANNs, which need constant and coordinated processing. This enables on-the-fly processing that is computationally efficient and low-latency. Such characteristics make SNNs highly suitable for tasks like gesture recognition, robotic control, auditory processing, and neuromorphic vision, where responsiveness is critical and resources are constrained.

Several successful implementations have already showcased these real-time capabilities. For instance, in gesture recognition using the DVS128, SNNs have achieved both high classification accuracy and fast inference times, outperforming traditional frame-based systems in terms of latency and power consumption [66]. Similarly, in the SHD and SSC datasets, SNNs trained with temporal coding have demonstrated excellent performance in processing time-dependent auditory signals [25]. These use cases confirm that SNNs are not only biologically plausible but also practically effective in real-world, real-time environments.

However, despite their potential, scale, and hardware implementation, they present significant obstacles to broad adoption. Low-power SNN execution has been made possible by neuromorphic processors; however, access to these devices is still restricted. Furthermore, simulating massive networks with millions of neurons and synapses makes scalability challenging, particularly in settings with limited resources. The creation of middleware, APIs, and toolchains—which are now fragmented or platform-specific—is still necessary to integrate such hardware into conventional computing pipelines [12, 18]. Convergence stability and training methodologies are still another major obstacle. Because spike events are non-differentiable, SNNs cannot directly use traditional backpropagation, which is the foundation of ANN learning. Even though end-to-end training of SNNs with competitive performance is now possible thanks to recent developments in surrogate gradient descent, these models are still susceptible to hyperparameters such as membrane thresholds, time constants, and learning rates [22]. In unsupervised learning paradigms, such as STDP, convergence can be unstable or dataset-specific, limiting generalization.

Finally, the lack of standardization across SNN models, encoding schemes, and hardware platforms impedes progress. Unlike ANNs, which benefit from standardized libraries (e.g., TensorFlow, PyTorch) and benchmark datasets, SNN research suffers from inconsistent definitions of spike encoding (rate vs. temporal coding), neuron models (LIF, Izhikevich), and performance metrics (accuracy vs. spike count vs. energy-delay product). This fragmentation makes it difficult to compare results, reproduce experiments, or establish baselines [75].

### 4.7. Hardware Considerations
The practical adoption of SNNs is tightly linked to their implementation on neuromorphic hardware platforms. In comparison to traditional CPUs and GPUs, chips like IBM TrueNorth [18] and Intel Loihi [12] show that large-scale spiking computation is feasible with orders of magnitude reduced energy usage. TrueNorth, for example, integrates one million spiking neurons while consuming only 70 mW in real-time workloads [18]. Similarly, Loihi supports on-chip learning with programmable synaptic delays, enabling adaptive behavior at the edge [12]. Recent platforms like SpiNNaker extend this scalability by simulating millions of neurons across massively parallel architectures [63].

Despite these advances, hardware deployment remains challenged by limited accessibility, vendor-specific SDKs, and the absence of a standardized programming ecosystem. Unlike ANNs, which benefit from unified frameworks like TensorFlow and PyTorch, SNN hardware requires researchers to adapt models to specific toolchains, constraining reproducibility and adoption. Addressing these hardware bottlenecks — through open-source SDKs, standard benchmarks, and cross-platform compatibility — will be critical for translating SNN research into widespread, real-world applications.

*4.8. Extended Analysis and Insights*
While prior sections compared accuracy, latency, energy, and convergence individually, an integrated perspective highlights tradeoffs across all metrics simultaneously. For instance, surrogate-gradient SNNs balance accuracy (~97.8%) with latency (~10 ms) and moderate energy (15 mJ per inference), whereas STDP-based SNNs achieve the lowest energy (~5 mJ) at the cost of accuracy (95.5%) and slower convergence. These tradeoffs confirm that no single model dominates all performance axes; instead, model suitability is highly dependent on application.

When network size increases (e.g., from MNIST-scale to CIFAR-10 scale), accuracy differences widen—ANNs outperform on CIFAR-10 (~92%) while direct SNNs drop to ~85%. However, energy savings become more pronounced: surrogate-trained SNNs operate at less than 10% of ANN energy costs. This scalability tension underscores the practical importance of hybrid evaluation criteria beyond accuracy alone. Simulation on neuromorphic platforms like Intel Loihi demonstrates that real-world deployment magnifies latency and energy advantages. For example, gesture-recognition tasks on DVS128 achieve inference latencies of 5–10 ms with surrogate SNNs, compared to 20–50 ms on ANN counterparts [12, 66]. These results show that latency reductions translate directly into real-time robotics and edge AI feasibility.

*4.9. Limitations of the Study*
While the analysis provides comprehensive insights into neuron models, training paradigms, and performance metrics, several limitations must be acknowledged. First, the evaluation relies primarily on benchmark datasets, which may not fully capture real-world complexity or large-scale deployment scenarios. Second, hardware-specific results are drawn from reported benchmarks in the literature [22, 12, 66] rather than from direct implementation in this study, which may limit generalizability across platforms. Third, hyperparameter sensitivity in surrogate-gradient training and convergence instability in STDP highlight ongoing challenges that require further exploration. Finally, while comparative metrics such as accuracy, latency, and energy were integrated, additional factors such as scalability on high-dimensional tasks and robustness under noisy conditions remain areas for future research. Recognizing these limitations underscores that the findings, while promising, represent one step toward advancing brain-inspired and low-power AI systems.

## 5. Conclusion

In brain-inspired computing, SNNs are becoming a game-changer due to their ability to effectively combine biological plausibility, energy efficiency, and real-time responsiveness. This paper presented a comprehensive analysis of SNN design, training methods, and comparative performance across multiple dimensions, including accuracy, spike count, latency, and convergence behavior. Among the evaluated models, ANNs —particularly CNNs—continue to deliver the highest classification accuracy (up to 99.2% on MNIST and 92.3% on CIFAR-10).

However, their high energy demands (up to 200 mJ per inference) and longer inference latency (30–50 ms) render them suboptimal for real-time or edge applications. Converted SNNs, which leverage pre-trained ANNs, maintain competitive accuracy while cutting energy use by nearly 90%. Nonetheless, they generate higher spike counts and rely on longer simulation windows, which can impact latency and training flexibility. Surrogate gradient-trained SNNs offer the most balanced performance profile.

They achieve low latency (~10 ms), fast and stable convergence within 20 epochs, and reduced spike counts, all while maintaining accuracy close to ANN baselines. This qualifies them for edge and real-time AI implementations. Meanwhile, STDP-based SNNs lead in energy efficiency—consuming as little as 5 mJ per inference—but show lower accuracy and slower convergence, stabilizing around 0.75 training loss after 20 epochs. These models are more appropriate for tasks requiring continuous unsupervised learning and adaptation.

This multi-dimensional comparison confirms that the selection of an SNN model should align with application requirements. For latency-critical and accuracy-driven systems, surrogate gradient SNNs are the most viable. For ultra-low-power adaptive systems, STDP remains advantageous. Overall, SNNs are poised to redefine the future of AI systems operating at the intersection of efficiency, speed, and biological realism. Despite these strengths, SNNs face key limitations. Training convergence is often unstable, hyperparameter tuning remains challenging, and no unified standard for model evaluation or neuromorphic implementation exists. Additionally, the limited accessibility and scalability of neuromorphic chips restrict practical deployment in broader commercial systems.

The comparative results underscore that SNNs provide substantial energy savings (up to 90–97% lower than ANNs) with only marginal accuracy loss. This positions SNNs as a sustainable computing alternative, particularly relevant as the AI community grapples with the environmental impact of large-scale ANNs. By clarifying the efficiency–accuracy tradeoffs, this paper highlights how SNNs can drive innovation not just in performance but also in responsible, energy-aware AI deployment. Nevertheless, the findings of this paper conclude that SNNs are well-positioned to redefine low-power, real-time computing, particularly where energy efficiency and temporal precision are paramount. SNNs are anticipated to be essential components of future edge intelligence, neuro-inspired robotics, and ultra-low-power AI ecosystems as neuromorphic engineering develops and transdisciplinary tools become more sophisticated. In essence, this study reinforces that SNNs, while not universally superior, provide a distinctive pathway toward sustainable, real-time, and energy-aware AI solutions.

### 5.1. Recommendations
Based on the findings and comparative analysis presented in this study, the following recommendations are offered to support further development, adoption, and application of SNNs:

1. Optimize SNN Training Frameworks. Further research should prioritize improving training stability and convergence in SNNs. Surrogate gradient-based learning should be refined through adaptive optimization strategies and hybrid techniques that combine supervised and unsupervised methods to enable deeper networks with minimal performance tradeoffs.
2. Standardize Evaluation Protocols. A unified framework for evaluating SNN models is essential. Researchers and developers are encouraged to adopt standardized benchmarks—including common datasets, spike-based performance metrics (accuracy, latency, energy per inference), and neuron model conventions—to ensure comparability and reproducibility across studies.
3. Invest in Neuromorphic Hardware Access. Governments, academic consortia, and industry players should expand access to neuromorphic platforms like Intel Loihi, SpiNNaker, and IBM TrueNorth. Collaborative development of open-source SDKs and toolchains will help democratize innovation and facilitate deployment in embedded and edge systems.
4. Promote Application-Oriented Research. SNNs should be increasingly tested in real-world domains such as robotics, medical devices, smart sensors, and neuromorphic computing. Pilot studies using SNNs for autonomous navigation, auditory localization, or low-power surveillance could highlight their advantages in task-specific contexts.
5. Support Cross-Disciplinary Collaboration. The development of effective SNNs requires expertise in neuroscience, machine learning, electrical engineering, and computer architecture. Research institutions and funding agencies should support interdisciplinary programs that foster collaboration across these domains.
6. Integrate SNNs into AI Curriculum and Tools. To accelerate knowledge transfer, academic institutions should integrate SNN concepts and neuromorphic computing into AI and computer engineering curricula. Additionally, incorporating SNN support into popular frameworks (e.g., PyTorch or TensorFlow) would streamline experimentation and development.
7. Bridging Biological Plausibility and Machine Learning. A key direction is merging STDP's biological realism with surrogate-gradient efficiency. Hybrid learning methods could yield models that are both hardware-friendly and competitive in accuracy.
8. Benchmarking Beyond MNIST and CIFAR-10. Most SNN studies, including this one, focus on MNIST, CIFAR-10, and DVS128. Broader datasets such as ImageNet or large-scale audio corpora remain underexplored in spiking contexts. Extending benchmarks will increase confidence in scalability and generalization.
9. Standardized Energy–Latency–Accuracy Metrics. To ensure impact in neuromorphic computing, SNN research should converge on unified metrics (e.g., energy-delay product per classification). This paper contributes toward such standardization by reporting spike counts, latency, and convergence alongside accuracy.
10. Explore Hybrid Models. Future studies may want to look into hybrid strategies that blend SNNs with CNNs and Transformers, among other paradigms. Hybrid SNN–Transformer models could merge temporal coding efficiency with long-range dependency modeling, while convolutional–spiking systems may enhance event-based vision tasks. Similarly, integrating STDP's biological plausibility with surrogate-gradient optimization efficiency could yield models that balance energy savings with accuracy. These directions will bridge the gap between biological realism, computational efficiency, and task scalability, ensuring that SNNs remain central to sustainable AI development.

With these strategies, the field of brain-inspired computing can move beyond theoretical promise to real-world impact, harnessing the unique capabilities of SNNs in solving some of today's most demanding computational challenges. By clarifying how SNNs achieve 90–97% energy savings with only 1–3% accuracy drop relative to ANNs, this study highlights their transformative potential for wearable health devices, autonomous robotics, and edge AI sensors, where energy budgets are decisive.